\documentclass[10pt,twocolumn,letterpaper]{article}

\usepackage{cvpr}
\usepackage{times}
\usepackage{epsfig}
\usepackage{graphicx}
\usepackage{amsmath}
\usepackage{amssymb}
\usepackage{algorithm}
\usepackage{algorithmic}
\usepackage{subcaption}
\usepackage{CJKutf8}

\usepackage[breaklinks=true,bookmarks=false]{hyperref}

\cvprfinalcopy 

\def\cvprPaperID{****} 

\begin{document}
\begin{CJK}{UTF8}{gkai}

\title{SuperChat:~Dialogue Generation by Transfer Learning from Vision to Language using Two-dimensional Word Embedding and Pretrained ImageNet CNN Models}

\author{Baohua Sun,  ~Lin Yang, ~Michael Lin,  ~Charles Young, ~Jason Dong, ~Wenhan Zhang, ~Patrick Dong\\
Gyrfalcon Technology Inc.\\
1900 McCarthy Blvd. Milpitas, CA 95035\\
{\tt\small \{baohua.sun,lin.yang\}@gyrfalcontech.com}
}

\maketitle

\begin{abstract}
The recent work of Super Characters method using two-dimensional word embedding achieved state-of-the-art results in text classification tasks, showcasing the promise of this new approach. This paper borrows the idea of Super Characters method and two-dimensional embedding, and proposes a method of generating conversational response for open domain dialogues. The experimental results on a public dataset shows that the proposed SuperChat method generates high quality responses. An interactive demo is ready to show at the workshop. 
\end{abstract}

\section{Introduction}
\begin{figure*}[t!]
\begin{center}
  \includegraphics[width=0.95\linewidth]{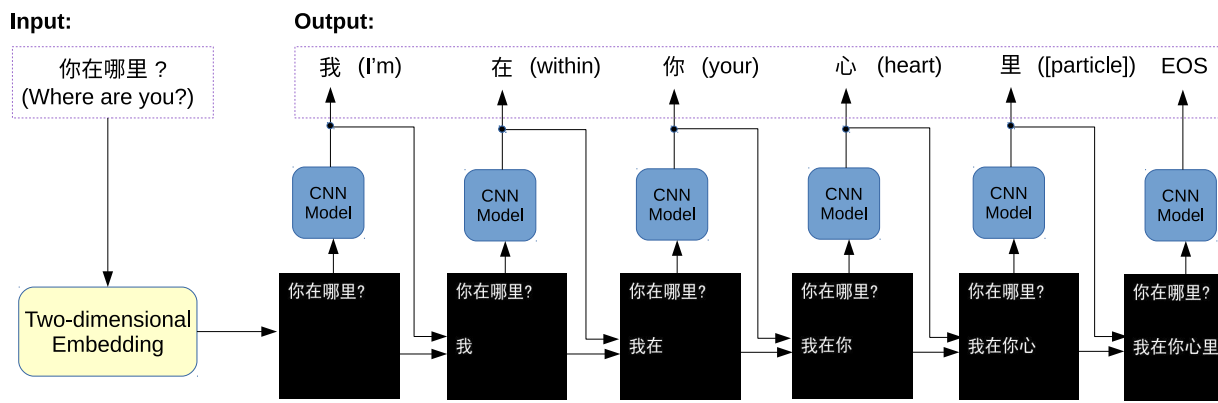}
\end{center}
\caption{SuperChat method illustration. The input Chinese sentence means ``Wher are you?" in English, and the output given by the proposed SuperChat method is ``I am within your heart". }
  \label{SuperChatIllustration}
\end{figure*}

Dialogue systems are important to enable machine to communicate with human through natural language. Given an input sentence, the dialogue system outputs the response sentence in a natural way which reads like human-talking. Previous work adopts an encoder-decoder architecture~\cite{sutskever2014sequence}, and also the improved architectures with attention scheme added~\cite{bahdanau2014neural,vaswani2017attention,wu2016google}. In architectures with attention, the input sentence are encoded into vectors first, and then the encoded vectors are weighted by the attention score to get the context vector. The concatenation of the context vector and the previous output vector of the decoder, is fed into the decoder to predict the next words iteratively. Generally, the encoded vectors, the context vector, and the decoder output vector are all one-dimensional embedding, i.e. an array of real-valued numbers. The models used in decoder and encoder usually adopt RNN networks, such as bidirectional GRU~\cite{bahdanau2014neural,gao2018generating}, LSTM~\cite{luong2015effective}, and bidirectional LSTM~\cite{wu2016google}. However, the time complexity of the encoding part is very expensive. 

The recent work of Super Characters method~\cite{sun2018super} has obtained state-of-the-art result for text classification on benchmark datasets in different languages, including English, Chinese, Japanese, and Korean. The Super Characters method is a two-step method. In the first step, the characters of the input text are drawn onto a blank image. Each character is represented by the two-dimensional embedding, i.e. an matrix of real-valued numbers. And the resulting image is called a Super Characters image. In the second step, Super Characters images are fed into a two-dimensional CNN models for classification. Examples of two-dimensional CNN models are used in Computer Vison (CV) tasks, such as VGG~\cite{simonyan2014very}, ResNet~\cite{he2016deep}, SE-net~\cite{hu2018squeeze} and etc. in ImageNet~\cite{imagenet_cvpr09}. The follow-up works using the two-dimensional word embedding also show the effectiveness of this method in other applications. The SEW~\cite{sun2019squared} method extends the Super Characters method to be applied in Latin languages such as English. And the SuperTML method~\cite{sun2019supertml} applies the two-dimensional word embedding to structured tabular data machine learning. 

In this paper, we propose the SuperChat method for dialogue generation using the two-dimensional embedding. It has no encoding phase, but only has the decoding phase. The decoder is fine-tuned from the pretrained two-dimensional CNN models in the ImageNet competition. For each iteration of the decoding, the image of text through two-dimensional embedding of both the input sentence and the partial response sentence is directly fed into the decoder, without any compression into a concatenated vector as done in the previous work. 

\section{The Proposed SuperChat Method}
The proposed SuperChat method is motivated by the two-dimensional embedding used in the Super Characters method. If the Super Characters method could keep the same good performance when the number of classes in the text classification problem becomes even larger, e.g. the size of dialogue vocabulary, then the Super Characters method should be able to address the task of conversational dialogue generation. This can be done by treating the input sentence and the partial response sentence as one combined text input. 

Figure~\ref{SuperChatIllustration} illustrates the proposed SuperChat method. The response sentence is predicted sequentially by predicting the next response word in multiple iterations. During each iteration, the input sentence and the current partial response sentence are embedded into an image through two-dimentional embedding. The resulting image is called as a SuperChat image. And then this SuperChat image is fed into a CNN model to predict the next response word. In each SuperChat image, the upper portion corresponses to the input sentence, and the lower portion corresponses to the partial response sentence. At the beggining of the iteration, the partial response sentence is initiallized as null. The prediction of the first response word is based on the SuperChat image with only the input sentence embedded, and then the predicted word is added to the current partial response sentence. This iteration continues until End Of Sentence (EOS) appeared. Then, the final output would be a concatenation of the sequential output. 

The CNN model used in this method is fine-tuned from pre-trained ImageNet models to predict the next response word with the generated SuperChat image as input. It can be trained end-to-end using large dialogue corpus. Thus the problem of predicting the next response word in dialogue generation is converted into an image classification problem.  

The training data is generated by labeling each SuperChat image as an example of the class indicated by its next response word. EOS is labeled to the SuperChat image if the response sentence is finished. 

The cut-length of sentences is high-related to the font size of each character. For fixed image size, the larger cut-length means smaller font size for each character, and vice versa. On the one hand, we want to cover long sentences, which means the cut-length should be big, so there will be variety in both the input dialogue and the response dialogue. On the other hand, if we set the cutlength too big, the font size of each character will be small, and there could be large blank area for short sentences, which is a waste of the space on the image. The cut-length should be configured according to the sentence length distribution.

It should be also emphasized that the split of the image into input and response part could be not even. Depending on the statistics of the training data, maybe larger or smaller size could be assigned to response and input text. Also, the font size for each part does not need to be the same.

Although the examples used in Figure \ref{SuperChatIllustration} is illustrated with Chinese sentences, however, it can be also applied to other languages. For example, Asian languages such as Japanese and Korean, which has the same square shaped characters as in Chinese. For Latin languages where words may have variant length, SEW method~\cite{sun2019squared} could be used to convert the Latin languages also into the squared shape before applying the SuperChat method to generate the dialogue response.

Beam search~\cite{freitag2017beam} could be also used. In that case, instead of hard prediction for the first character, a soft prediction will be used to output all the possible sentences and one of the best will be selected as the final output. 

\section{Experiments}
The dataset used is Simsimi\footnote{https://github.com/fate233/dgk\_lost\_conv/blob/master/results\\ /xiaohuangji50w\_nofenci.conv.zip}. This is a Chinese chitchat database. This data set contains 454,561 dialogue pairs. There are totally 5,523 characters used in the response sentences, of which 4996 are characters with frequency less than 1000. The top five frequency characters are ``,", ``我" (I), ``你" (you), ``的" (of), and ``是" (is).

Based on the distribution of the sentence length, we set cut length for input sentence at 18, and response cut length also at 18. So, altogether we have 36 characters within one SuperChat image, which could be a layout of 6 rows by 6 columns of characters. The input sentence takes the upper 3 rows, and the response sentence takes the lower 3 rows. For simplicity, we removed all the emoticons in the data set.  In order to get enough samples for training, only characters whose frequency is not less than 1000 appearances are selected in the list of characters to predict. After this filtering, the remaining set is composed by the sentences with both input and response sentence length less than 18 characters, and all its characters in the list of the 528 frequent characters (including EOS). The resulting set is 178,192 pairs of dialogues, and a total of 989,087 SuperCharacter images are generated.

We set our image size at three channels of 224x224 grey image, in order to use the pretrained models on ImageNet. We also added a margin area for the four edges in the SuperChat image, which means the first character will not start from the pixel location of $(0,0)$, but from $(m,m)$ instead. Here $m$ is the we set for the four edges. In this experiment, we set $m=16$, which results in the remaining area is the square of $224-16$x$2 = 192$ pixels. If we set same font size for both input and response sentence, it results in a font size of $192/6 = 32$ pixels. That means, each character takes an area of $32$x$32$ pixels. The fonts used is the ``simhei". 

\subsection{Model Training}
\begin{figure}[h]
\begin{center}
  \includegraphics[width=0.95\linewidth]{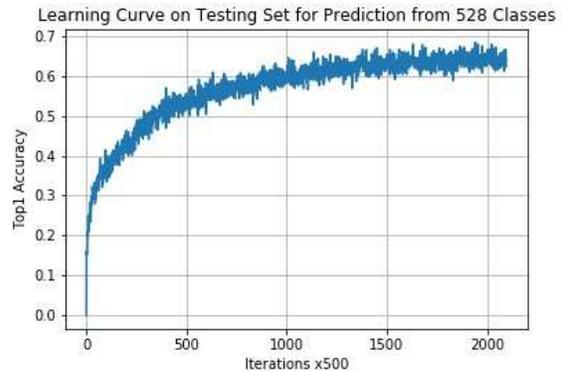}
\end{center}
\caption{Learning Curve on Testing Data}
  \label{LearningCurve}
\end{figure}

For each character, we split its labeled data into 75\% for training and 25\% for testing. Resulting in 739,289 training samples and 249,798 testing samples.
SE-net-154 is used in this experiment. The pretrained model on ImageNet is used as initialization\footnote{https://github.com/hujie-frank/SENet}. 
This model is fine-tuned on the generated SuperChat images just as an image classification task. The fine-tuning is done by simply modifying the last layer to 528, which is the size of the subset of the response vocabulary. The learning curve on the test data is shown in Figure~\ref{LearningCurve}~.

We can see that at the begining of the training, the curve climbs up quickly, and after it achieved 60\%, the slope goes slowly and almost saturate to 64\%. The x-axis is in unit of every 500 iterations, so totally it is one million iterations, with batch size of 5, wich accounts to about 7 epochs over the training data.

\subsection{Sample Response Sentences}
Table~\ref{SampleResponseSentences} are sample response sentences output by the SuperChat method.
We can see the responses follow the grammar rules, and the style of the response sentences are funny and cute, as learnt from the training data.
\begin{table*}[t]
\caption{Sample response sentences generated by the SuperChat method on the Simsimi data set.}
\label{SampleResponseSentences}
\begin{center}
\begin{tabular}{|c|l|l|}
\hline 
\bf Sample ID&\bf Input Sentence&\bf Response Sentence
\\ \hline 
1&``你有男朋友吗?"&``没有,我只有老公" \\
~&``Do you have boyfriend?"&``No,I only have a husband" \\
\hline 
2&``哼!你不想我我就不和你玩"&``我很好玩的跟我玩嘛跟我玩嘛" \\
~&``Hmm! I won't play with you if you didn't miss me"&``I am funny, please play with me, please" \\
\hline 
3&``我看到你第一眼,就喜欢你了"&``我也喜欢你" \\
~&``The first time I saw you, I liked you"&``I like you too" \\
\hline 
4&``是什么意思"&``你自己百度去" \\
~&``What does that mean"&``Go Baidu it yourself" \\
\hline 
5&``我去睡觉了"&``晚安，好梦，明天叫你起床哦" \\
~&``I am going sleep now"&``Good dream, wake you up tomorrow" \\
\hline 
6&``你什么时候睡觉"&``我等你睡了再睡" \\
~&``When will you sleep"&``I will wait untill after you sleep" \\
\hline 
7&``无聊中"&``我陪你聊天呗" \\
~&``I am bored"&``Let's chitchat" \\
\hline 
8&``我好难过,哭的都好累"&``不哭不哭，哭了不好看了" \\
~&``I am sad, and cried a lot"&``Don't cry, not looking good when crying" \\
\hline 
9&``问你家哪的"&``我是火星人" \\
~&``Where are you from"&``I am from Mars" \\
\hline 
10&``对啊你吃什么"&``你喜欢吃什么，我就喜欢吃什么" \\
~&``BTW what do you eat"&``I eat whatever you like to eat" \\
\hline 
11&``为你付出那种伤心你永远不了解"&``我不会骗你。现在不会、以后也不会。" \\
~&``You won't know my heartbroken for you"&``I won't lie to you. Not now, neither future." \\
\hline 
12&``那你是美女不"&``我是美女！" \\
~&``So you are a beauty"&``I am a beauty!" \\
\hline 
13&``睡了吗"&``没有啊，我在等你呢。" \\
~&``Are you asleep"&``Nop, I am waiting for you." \\
\hline 
14&``你会打鸣不"&``喔喔喔！" \\
~&``Can you do rooster crowing"&``Cockadoodledoo!" \\
\hline 
15&``你累么"&``你累吗？累了早点睡啦。乖乖" \\
~&``Are you tied"&``You tired? Then go sleep. Honey" \\
\hline 
16&``我想她了,怎么办"&``去找她啊！" \\
~&``I miss her, what should I do"&``Go see her!" \\
\hline 
17&``你什么都懂噢"&``我不知道,但有些东西我刚好知道" \\
~&``You seem to know everything"&``I don't konw, but it happens I know something" \\
\hline 
18&``我昨天发状态找你。你都不给回"&``我错了" \\
~&``I sent message to you. But you didn't reply"&``I apologize" \\
\hline 
19&``讨厌"&``我讨厌一切讨厌我的人" \\
~&``Dislike you"&``I dislike anyone who dislikes me" \\
\hline 
20&``猜猜我是什么星座"&``双子座" \\
~&``Guess what is my constellation"&``Gemini" \\
\hline 
\end{tabular}
\end{center}
\end{table*}

\section{Conclusion}
In this paper, we propose the SuperChat method for dialogue response generation. It has no encoding, but only decodes the two-dimensional embedding of the input sentence and partial response sentence to predict the next response word iteratively. The pretrained two-dimensional CNN model is fine-tuned with the generated SuperChat images. The experimental results shows high quality response. With low-power CNN accelerators becoming widely available~\cite{sun2018ultra,sun2018mram}, the proposed SuperChat method could be used for the on-device chatbot system which generates the dialogue at the edge. An interactive demonstration is to show at the workshop.
{\small
\bibliographystyle{ieee_fullname}
\bibliography{egbib}
}
\end{CJK}
\end{document}